\documentclass[preprint,12pt]{elsarticle}

\usepackage{graphicx}
\graphicspath{{./Figures/}}
\usepackage{subfigure}

\usepackage{amssymb}
\usepackage{lineno}
\usepackage{url}
\usepackage{booktabs}
\usepackage{siunitx}
\journal{Acta Tropica}

\begin{document}

\begin{frontmatter}


\title{Modeling Dengue Vector Population Using Remotely Sensed Data and Machine Learning}



\author[UNC]{Juan M. Scavuzzo}
\author[UNC]{Francisco Trucco}
\author[MundoSano]{Manuel Espinosa}
\author[CONAE]{Carolina B. Tauro}
\author[MundoSano]{Marcelo Abril}
\author[CONAE]{Carlos M. Scavuzzo}
\author[Angolas]{Alejandro C. Frery}

\address[UNC]{Facultad de Marem\'atica, Atronom\'ia, F\'isica y Computaci\'on, Universidad Nacional de C\'ordoba}
\address[CONAE]{Instituto de Altos Estudios Espaciales Mario Gulich, Universidad Nacional de C\'ordoba-Comisi\'on Nacional de Actividades Espaciales}
\address[MundoSano]{Fundaci\'on Mundo Sano, Buenos Aires, Argentina}
\address[Angolas]{Universidade Federal de Alagoas, Brazil}

\begin{abstract}
Mosquitoes are vectors of many human diseases. 
In particular, \textit{Aedes \ae gypti} (Linnaeus) is the main vector for Chikungunya, Dengue, and Zika viruses in Latin America and it represents a global threat. 
Public health policies that aim at combating this vector require dependable and timely information, which is usually expensive to obtain with field campaigns. 
For this reason, several efforts have been done to use remote sensing due to its reduced cost. 
The present work includes the temporal modeling of the oviposition activity (measured weekly on \num{50} ovitraps in a north Argentinean city) of \textit{Aedes \ae gypti} (Linnaeus), based on time series of data extracted from operational earth observation satellite images. 
We use are NDVI, NDWI,  LST night, LST day and TRMM-GPM rain from \num{2012} to \num{2016} as predictive variables. 
In contrast to previous works which use linear models, we employ Machine Learning techniques using completely accessible open source toolkits. 
These models have the advantages of being non-parametric and capable of describing nonlinear relationships between variables. 
Specifically, in addition to two linear approaches, we assess a Support Vector Machine, an Artificial Neural Networks, a K-nearest neighbors and a Decision Tree Regressor. 
Considerations are made on parameter tuning and the validation and training approach. 
The results are compared to linear models used in previous works with similar data sets for generating temporal predictive models. 
These new tools perform better than linear approaches, in particular Nearest Neighbor Regression (KNNR) performs the best. % ARO: cambio aca.
These results provide better alternatives to be implemented operatively on the Argentine geospatial Risk system that is running since \num{2012}.  
\end{abstract}

% NOTA CARO: cambie  "KNNR perform the best"  por  "K-Nearest Neighbour Regression (KNNR) and Multilayer Perceptron"  que es lo que dice la conclusion.


\end{frontmatter}

%\linenumbers

\section{Introduction}
\label{S:1}

Machine Learning (ML) is an effective empirical approach for regressions and/or classification of nonlinear systems which may involve from a few to thousands of variables. 
The ML approach requires training data covering most of the system's parameter space. More often than not, a subset of these data is kept for validation. 
ML is ideal to address those problems in which our theoretical knowledge is still incomplete but for which we have a large number of observations. 
ML has been shown to be useful for a large number of applications in Geosciences for land, oceans and atmosphere, and in bio-geophysical information extraction algorithms \cite{Lary2009, Brown2008, Azamathulla2012, Zahabiyoun2013,  Madadi2015, Yi1996}.

Some of the most used ML algorithms in Geosciences and Remote Sensing (GRS) applications are Artificial Neural Networks (ANN), Support Vector Machines (SVM), Self-Organizing Maps (SOM), Decision Trees (DT), Random Forests, and Genetic Algorithms \cite{Lary2016}. 
Their application in GRS problems is relatively new and extremely promising \cite{Lary2016, Penia2014}. 
In particular, ANNs are widely used for classification but also for time series forecast \cite{Atkinson1997,Zhang2005,Foody2004}.  
In fact, an exploration in the bibliographic base Scopus returns more than \num{4000} publications that include ``remote sensing'' and ``neural network'', \num{311} of them in \num{2016}. 
Of this total \SI{45}{\percent} correspond to the area of ``Sciences of the Earth'', \SI{44}{\percent} to ``Computer Science'' and \SI{35}{\percent} to ``Engineering'', with China, the United States, Italy and India  the countries with the highest scientific production in the area \cite{Bose2017,Wang2016,JafariGoldarag2016}. 
None, to the best of the authors' knowledge, deals with Epidemiology, Remote Sensing and ML.

Mosquitoes are the most important vectors of human diseases.  
In particular, \textit{Aedes \ae gypti} (Linnaeus) is the main vector for Chikungunya, Dengue, and Zika viruses. 
This is a peridomestic mosquito that is bred preferably in artificial containers \cite{Powell2013}, \cite{Moncayo2004}. 
The incidence of Dengue has increased dramatically in the last decades, with a rising trend of outbreaks in South America in recent years, and Chikungunya and Zika are new threats spread by the same species of mosquito~\cite{WHO2015} , \cite{who2015b}, \cite{who2016}.  
The deployment of ovitraps is generally accepted as a valid method to provide useful data on the spatial and temporal distribution of \textit{Aedes \ae gypti} (Linnaeus), allowing a reasonable estimation of vector activity~\cite{Ritchie1084}.

Landscape Epidemiology~\cite{Ostefeld2005,Pavlovsky1966} promotes the notion that satellite data from earth observation and geospatial technologies are essential tools~\cite{Hay2000} to address vector borne epidemiological problems. 
Using these ideas, several interdisciplinary studies were produced in latinoamerica focused in generating spatial and temporal predictive risk models based on satellite derived environmental conditions  ~\cite{Parra2010,Douglas2010,moreno2014,arboleda2012}.   
In particular in Argentina there are  interesting experiences on this issue, for example~\cite{Rotela2007,Estallo2011,Espinosa2016b} deal with Dengue epidemics from the dynamic point of view, while~\cite{Porcasi2012} are concerned with the deployment of operational tools for its management.
At a global scope we can find interesting contributions ~\cite{Herbreteau2007,Kalluri2007,Buczak2012} with also some operatives experiences ~\cite{Bowman2016}  

Specifically, in a interinstitutional framework between the Argentinean National Space Agency (CONAE) and the Health Ministry of Argentina, there have been initiatives to model the temporal evolution of mosquito populations using environmental variables obtained from remote sensors. 
These works used series of a few years and are based on a small number of satellite variables \citep{Estallo2012,Estallo2016}.
In an effort to improve this, \cite{German2017} constructed models based on a large number of variables from various sensors for four years. 
All these works assumed multivariate linear models.

This work represent an improvement of that scenario.
We compare  Support Vector Machines, Artificial Neural Networks, K-nearest neighbors and Decision Tree Regressor in addition to two linear approaches. 
With this, we obtain an operational methodology which contributes to the Argentinean Dengue risk system currently in operation~\cite{Porcasi2012,Rotela2017}.

We explore, in contrast to previous ones, the ability of modeling and predicting oviposition without of the shelf” ML algorithms, i.e., with minimum parameter tuning, as provided by FLOSS -- Free/Libre Open Source Software. 
This promotes the assimilation of these techniques for the whole community that deals with similar problems.

\section{Materials}

\subsection{Study Area and Field Data}

\begin{figure}[hbt]
\centering
\includegraphics[width=0.7\textwidth]{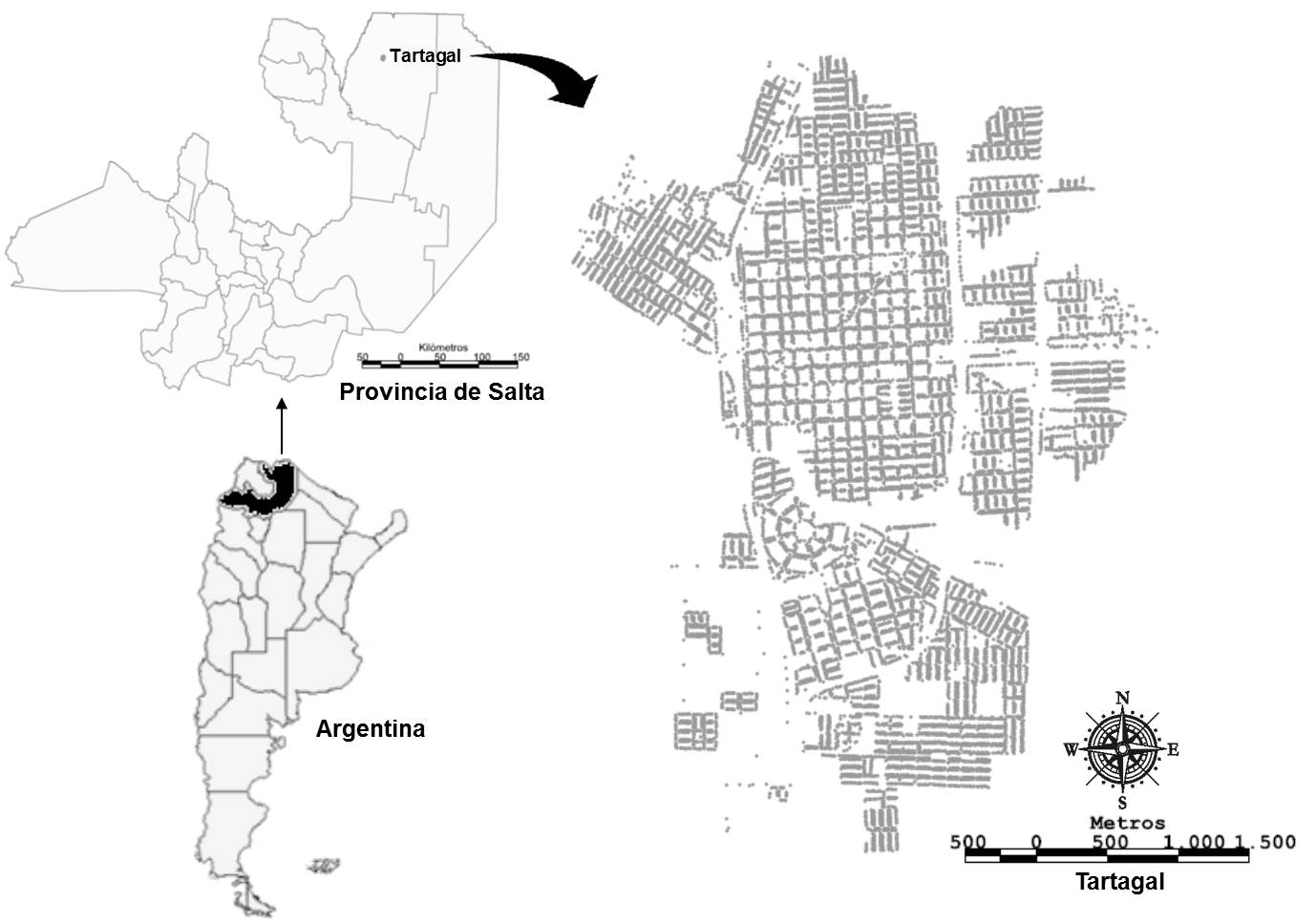}
\caption{Study area}\label{Fig:StudyArea}
\end{figure}

The study here presented was developed on Tartagal city (79,900 inhabitants) on the Northwest of Argentina (\ang{22;32;}~S, \ang{63;49;}~W, \SI{450}{\meter} above the sea level), in Salta Province. The site is between \num{50} and \SI{100}{\kilo\meter} from the Argentinean-Bolivian border (Figure~\ref{Fig:StudyArea}). 
Tartagal is in a subtropical native forest environment surrounded by crops. 

The site has an average annual temperature of about \SI{23}{\degreeCelsius} (summer average maximum of \SI{39}{\degreeCelsius} and winter average minimum of \SI{9}{\degreeCelsius}). 
It has an annual precipitation of \SI{1100}{\milli\meter}, with a dry season (June to October). 
Tartagal, like several north-west Argentinean cities, has a cultural diversity based on the presence of autochthonous ethnic groups and immigrant population in addition to a migration movement from the bordering country Bolivia. 
These characteristics lead to peculiar cultural, social and economic profile behavior. 

The vector population is measured using the monitoring of oviposition activity. 
It is measured using ovitraps placed at randomly selected houses in the urban area of the City.  
The period of monitoring used in this study was from August \num{2012} until July \num{2016} over \num{50} houses. 
Two ovitraps were placed in each house: one inside and other outside in a shaded site at ground level in the backyard, following the WHO guidelines~\cite{WHO2015}. 
The ovitraps are \SI{1000}{\centi\meter\cubed} of black plastic cups containing \SI{250}{\milli\liter} of water without attracting infusion. 
We used only the external ovitraps data in this study because they correlate more with the satellite-derived environmental variables. 
The ovitraps are replaced weekly and eggs are counted on a laboratory according to the Egg Density Index~\cite{Gomes1998}. 
Then the weekly \textit{Aedes \ae gypti} (Linnaeus) oviposition activity is estimated by the sum of egg-catches on the external traps of the city.

\subsection{Environmental Variables}

Following the idea to build predictive models of vector population based on environmental variables derived from satellite, but with an operational perspective and based on previous studies, we obtain proxies of the vegetation, moisture, temperature and rain operationally available from MODIS and TRMM/GPM products. 

Global vegetation indexes provide consistent spatial and temporal products of vegetation canopy greenness, property of leaf area, chlorophyll and canopy structure. 
These indexes are derived from atmospherically-corrected reflectance in the red and near-infrared bands. 
In our case, we use the NDVI from the MODIS MOD13Q1 satellite product (composed of 16 days) with a \SI{250}{\meter} spatial resolution. 
The vegetation conditions are included because it is related also with the temperature, humidity and precipitation~\cite{Estallo2012,Hay1997}, relevant variables for the mosquito population evolution.

In addition, we include the Normalized Difference Water Index (NDWI), which is related to the liquid water and humidity content in both soil and vegetation.
It is calculated from the same MODIS product using Gao's definition~\cite{Gao1996} of NDWI from the bands provided by the MOD13Q1 product, corresponding to MIR and NIR reflectance $NDWI =  (\rho_{NIR} - \rho_{MIR}) / (\rho_{NIR}  + \rho_{MIR} )10^4$. 
MODIS products require the $10^4$ factor since they are stored, for computational economy, as integer numbers.

We also used Land Surface Temperature (LST) from MODIS because it is an approximation of the environmental temperature~\cite{Kalluri2007,Peres2004,Wan1999}. 
For this, the MOD11A2 satellite product was chosen. 
It has \SI{1}{\kilo\meter} spatial resolution and is an average of clear-sky LST's values during an \num{8}-day period. 
This product includes daytime and nighttime LST's representing, in some sense, the maximum and minimum temperatures~\cite{Wan2004}. 

Local precipitation is obtained from the Tropical Rainfall Measuring Mission (TRMM)~\cite{Kummerow1998}. 
This is a joint mission of NASA and the Japan Aerospace Exploration Agency launched in \num{1997} to study rainfall for weather and climate research. 
The satellite uses several instruments including radar, microwave imaging, and lightning sensors,  to detect rainfall. 
TRMM was out of fuel on \num{2014}, even though it continued providing data until June \num{2015}. 
After that, other products were published to assure continuity in the information based in a new space mission called GPM (\url{https://earthdata.nasa.gov/trmm-to-gpm}).

\begin{figure}[hbt]
\centering
\includegraphics[width=0.45\textwidth]{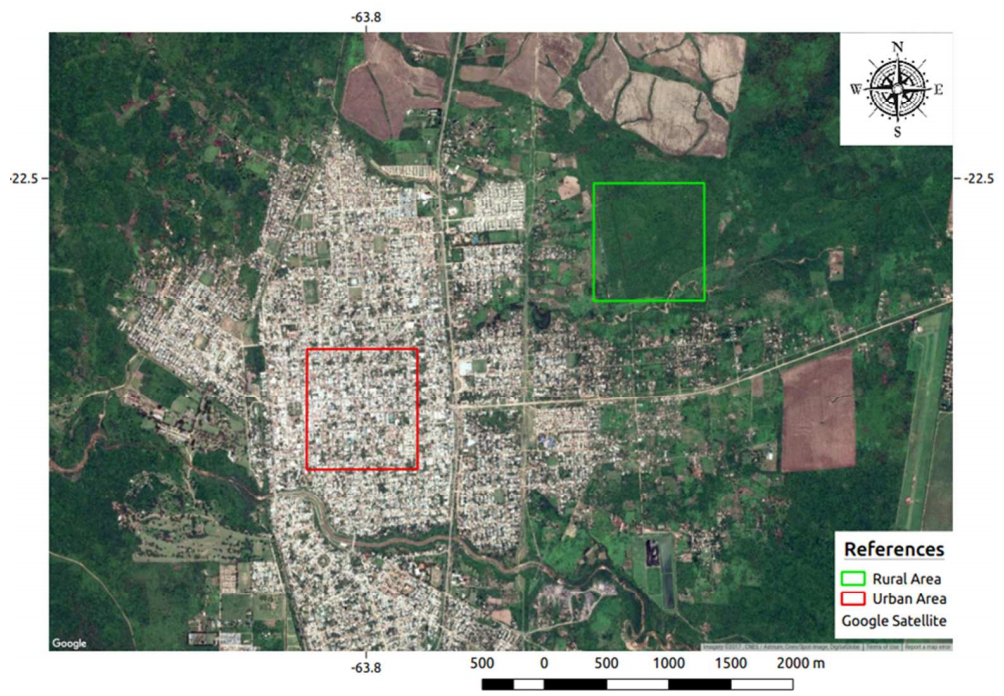}
\caption{urban and rural selected areas to extract the environmentals variables}\label{Fig:zonas}
\end{figure}

Two areas of \SI{85}{\hectare} were defined around the city and then the mean values, for all the satellite derived variables, were calculated. 
The first area is located within the city (Urban Area) and the second one encompasses the native vegetation surrounding the city (Rural Area) following the approach presented by \cite{Estallo2008,Estallo2014,German2017}. 
The choice was made under the hypothesis that selecting a zone outside the city would represent well the environmental conditions (NDWI, NDVI, and LST). 
%The same idea was used in previous studies in several cities. 
%These observations and the larval indexes are likely to be closely related. 
%In this way, the external or rural area should be randomly selected from areas of native vegetation close enough to the city, representing the natural environment conditions. 
In this specific case this rural region is selected in the north-east of the city. 
It has a similar altitude to the city and mostly native forest.  
It can be see in Figure~\ref{Fig:zonas}.   

\begin{figure}[hbt]
\centering
\includegraphics[width=0.45\textwidth]{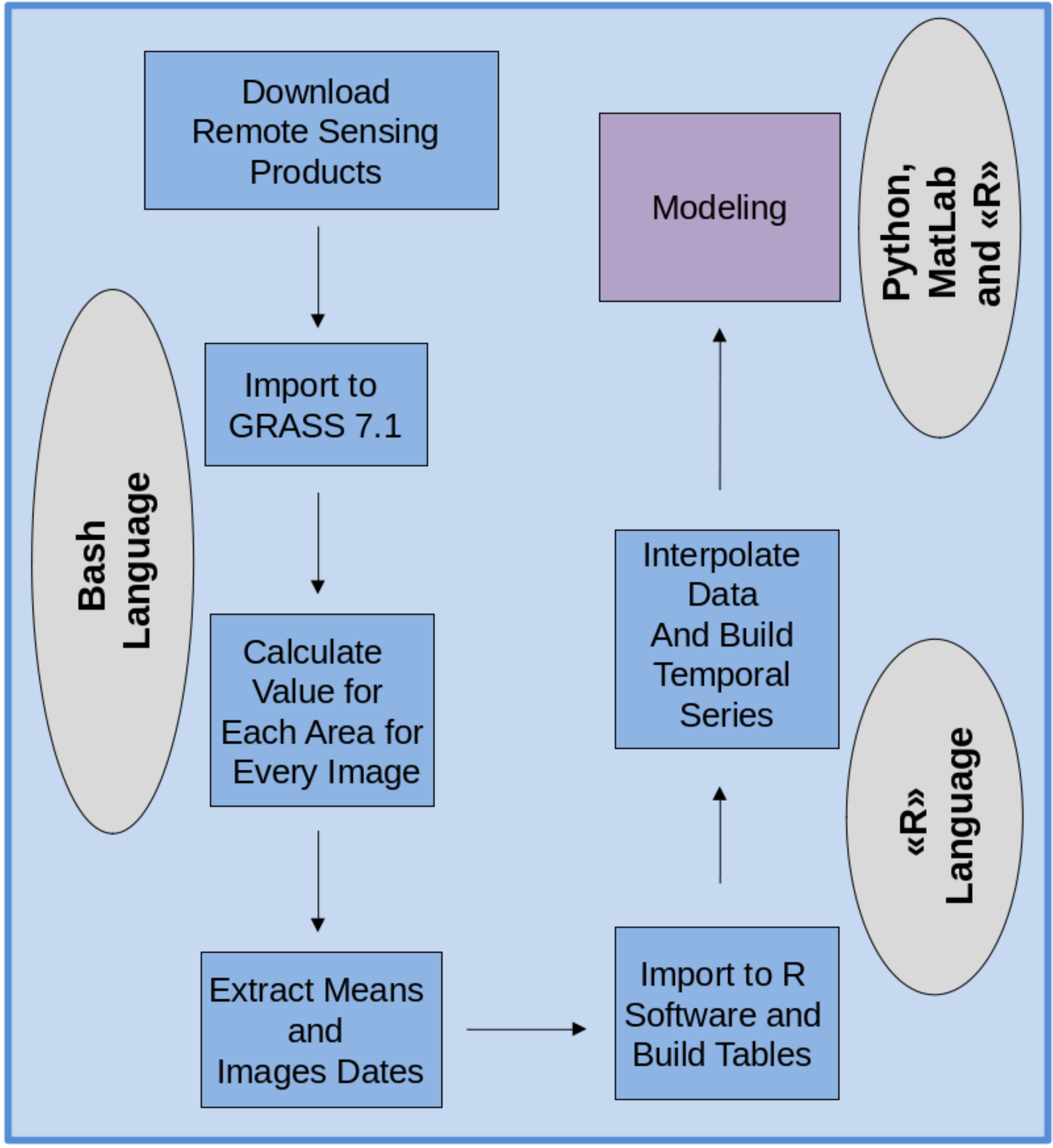}
\caption{Satellite products processing}\label{Fig:SatProdProcc}
\end{figure}

The procedure to build the temporal series of remote sensed variables is outlined in Figure~\ref{Fig:SatProdProcc}. 
The images were downloaded from NASA (\url{http://e4ftl01.cr.usgs.gov}) and imported into GRASS~7.1. 
The mean for each of the previously defined two areas was calculated for every date.  
All these average values and their dates were exported to a table in the R software, which is used to build the complete temporal series. 
The data were interpolated in order to obtain values for all the sampling dates (a value for every epidemiological week). 

All the variables are considered with three weeks lags from the original time series, to represent Non synchronous influences, corresponding to one, two and three time lapses.

The first step consisted in analyzing the forty environmental variables and eggs collected in each week by means of a correlation matrix and the $p$-values that measure their significance. 
This led to discarding thirty-five variables. 
Lagged variables were preferred because of their potential ability to forecast. 
The following variables were chosen:  NDVI rural lag 1, NDWI rural lag 1, LST day rural lag 3, LST night rural lag 1, and TRMM lag 3. 
All the variables are then normalised using z-scores. 

\begin{figure}[hbt]
\centering
\includegraphics[width=\linewidth]{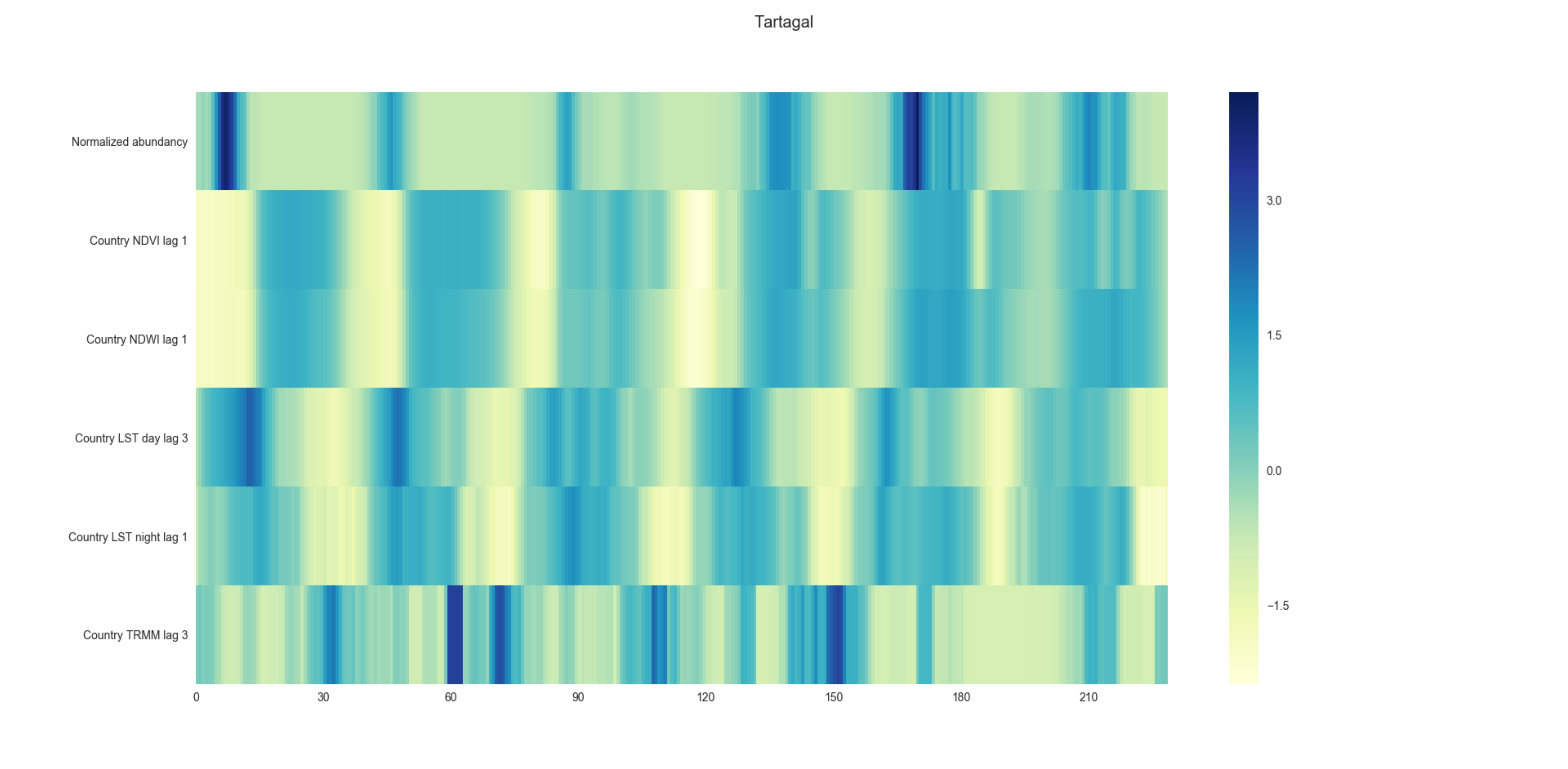}
\caption{Heatmap of temporal series of normalised z-score oviposition and  environmental variables. Time is in weeks}\label{Fig:Heatmat}
\end{figure}

Figure~\ref{Fig:Heatmat} presents the environmental variables along with the oviposition data as a heatmap. 
This format promotes the visualization of the temporal evolution, the correlation pattern between variables, and the lags effect.

\section{Modeling}

In order to model the oviposition as a function of time (weeks), we implemented two linear models (Simple and Ridge) and four non-linear models (Support Vector Machine, ANN multi-layer Perceptron, Decision Tree, and K-Nearest Neighbor). We used the same set of five environmental variables, which were described in previous section, for all those models.

%With the data described in previous section, we implemented two linear models (Simple and Ridge) and four non-linear models (Support Vector Machine, ANN multi-layer Perceptron, Decision Tree, K-Nearest Neighbor) to model the oviposition in each week. All the models use the same set of 5 environmental variables

In all the cases we generated the models with \SI{80}{\percent} of the dataset and retained the remaining \SI{20}{\percent} of the temporal series (almost one year) as an independent set to corroborate the temporal prediction capacity of the tools (we use the last \SI{20}{\percent} from our dataset). 
This splitting selection is the most used in the ML literature \citep{Cramer2017}. 

Cross validation~\cite{Picard1984,Cramer2017} was used in order to decrease the dependency of the evaluation results on a particular selection of training set and validation set pair. 
In particular, a time series split cross validation procedure was used to evaluate the models  \url{http://scikit-learn.org/stable/modules/cross_validation.html}. 
Other cross validation techniques like K-folds are not suitable for time series data, i.e., when the ordering of the data is relevant.

In the following we describe the techniques used to model the oviposition z-score as a function of the remotely sensed environmental variables. 
All the models were implemented using functions from the sklearn library, freely available in Python.

\subsection{Linear Regressions}

Previous experiences on the modeling of epidemiological applications using remotely sensed  environmental variables report good results with this approach~\cite{Andreo2009,Estallo2016,Ra2012}.
We used simple linear and ridge regressions, the latter with Tikhonov regularization with cross-validation. 
Note that Ridge regression is often referred to as ``weight decay'' in the ML literature.

\subsection{Nonlinear Models}

Nonlinear models are able to capture more complex functional relations among the data, at the expense of computational complexity and some burden on the user that has to fine tune more parameters than in linear models.
	
Typically, machine learning regression includes three steps: architecture, e.g. the number of layers and neurons in an artificial neural network or the number of neighbours in the K-Nearest Neighbor algorithm, the training-validation (where the coefficients are adjusted and the performance is evaluated), and then the use of the model with new data. These steps were implemented with functions available in the sklearn package already mentioned.

The configuration or selection of the optimal set of parameters in this kind of nonlinear models is a complex issue and could be handcrafted or obtained using semi automatic tools. 

We used the iRace (Iterated Racing for Automatic Algorithm Configuration) package \cite{Lopez2016} for automatic parameter tuning. This tool is an iterative procedure capable of automatically finding the most appropriate parameter configurations given the input data instances of the optimization problem. It is implemented in R and is freely available at \url{http://iridia.ulb.ac.be/irace/}. 

In order to avoid overfitting, a problem when dealing small data sets,  the tuning was performed automatically with data from a different city: Clorinda.

\subsubsection{Support Vector Regressor (SVR)}

Support Vector Machines are a class of supervised techniques that build either linear or nonlinear decision rules and regression models. We used the SVR from SVM module. This method implements Epsilon-Support Vector Regression, with penalty C = 0.887453, and RBF kernel coefficient gamma = 0.015561 as tuning parameters.

\subsubsection{Multilayer Perceptron (MLP)}

Neural Networks are built by a massive number of simple processing units highly interconnected. They can be trained to provide universal function approximators. We used the MLPRegressor method from the \verb|neural_network| module. This method implements the Multilayer Perceptron regressor by optimizing the squared loss by either LBFGS or stochastic gradient descent. We tuned the following parameters: alpha (the regularization quadratic term, set to 0.070921), three layers with three neurons each proved being a suitable and parsimonious architecture for our problem. The activation is done by the rectified linear unit function $f(x) = \max\{0, x\}$.

\subsubsection{k-Nearest Neighbour Regression (KNNR)}

We used the \verb|K-NeighborsRegressor| module.
This method infers a regression based on k-nearest neighbors. The target is predicted by local interpolation of the targets in the neighborhood in the training set. 
The original data are decomposed with principal components, and only the first five are used. 
The tuning parameters choices were four neighbors, uniform weight, Chebyshev metric and brute force.

\subsubsection{Decision Trees Regression (DTR)}

Decision Trees are classification rules built incrementally, from which a regression model can be learned. We used the \verb|DecisionTreeRegressor| method from the tree module. Again, we used PCA but retained only the two first components. The other parameters were the splitting rule (``best''), the maximum depth of the tree (three levels), and the minimum number of samples required to split an internal node (five).

The choice of numbers of PCA components for the two last methods was based on trial-and-error, seeking for the smallest subset that produced good results.

\section{Results}

%Figure~\ref{Fig:All} shows the z-scores of the observed data and those from the fitted models. 
%The latter follow the former, and in the following we analyze each technique.

%\begin{figure}[hbt]
%\centering
%\includegraphics[width=\linewidth]{All.pdf}
%\caption{Observed z-scores and all fitted data. Time is measured in weeks}\label{Fig:All}
%\end{figure}

\begin{figure}[hbt]
\centering
\includegraphics[width=\linewidth]{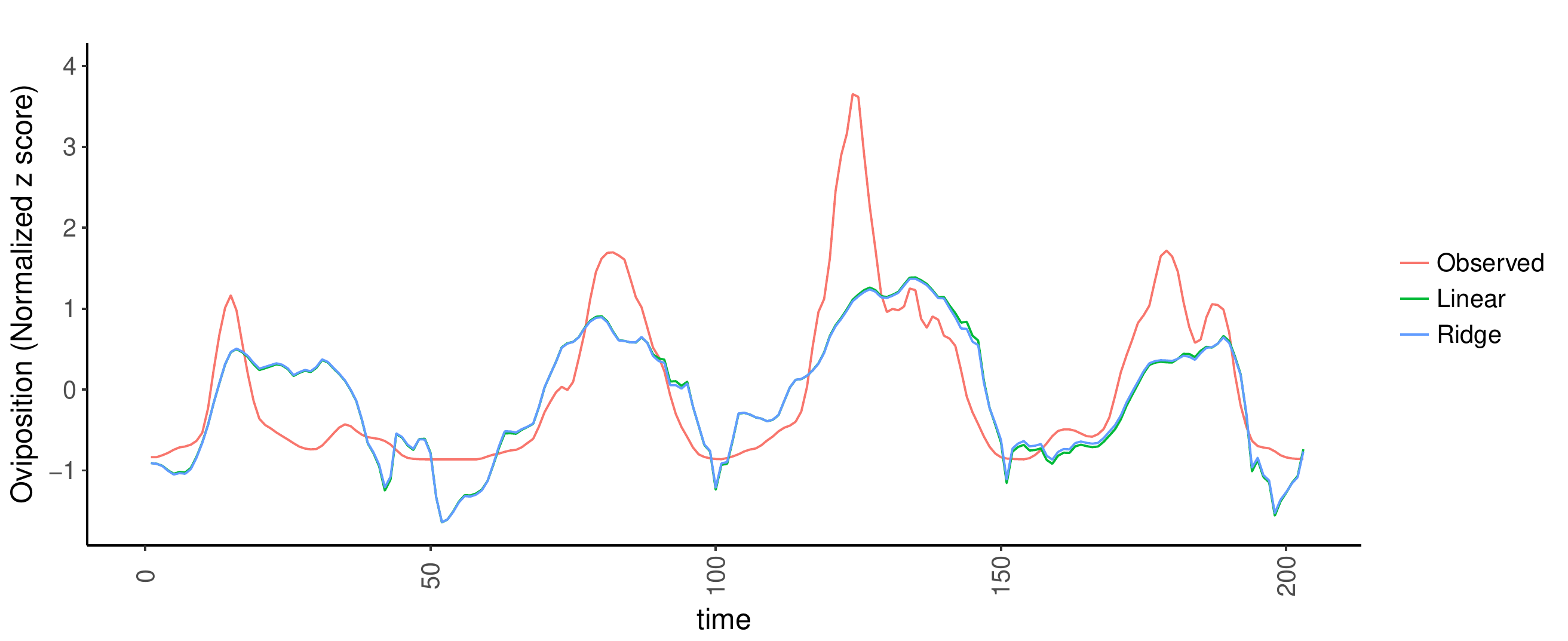}
\caption{Observed z-score, Linear and Ridge regressions}\label{Fig:LinearRidge}
\end{figure}

Figure~\ref{Fig:LinearRidge}  shows the results of the classical multivariate linear and Ridge models. 
These results are in conformity with previous studies. 
Both linear regressors produce very close results preventing, thus, the use of the latter due to its higher computational cost.

Linear regressors do not follow the peaks of the observed data, and tend to underestimate the smallest values.

\begin{figure}[hbt]
\centering
\includegraphics[width=\linewidth]{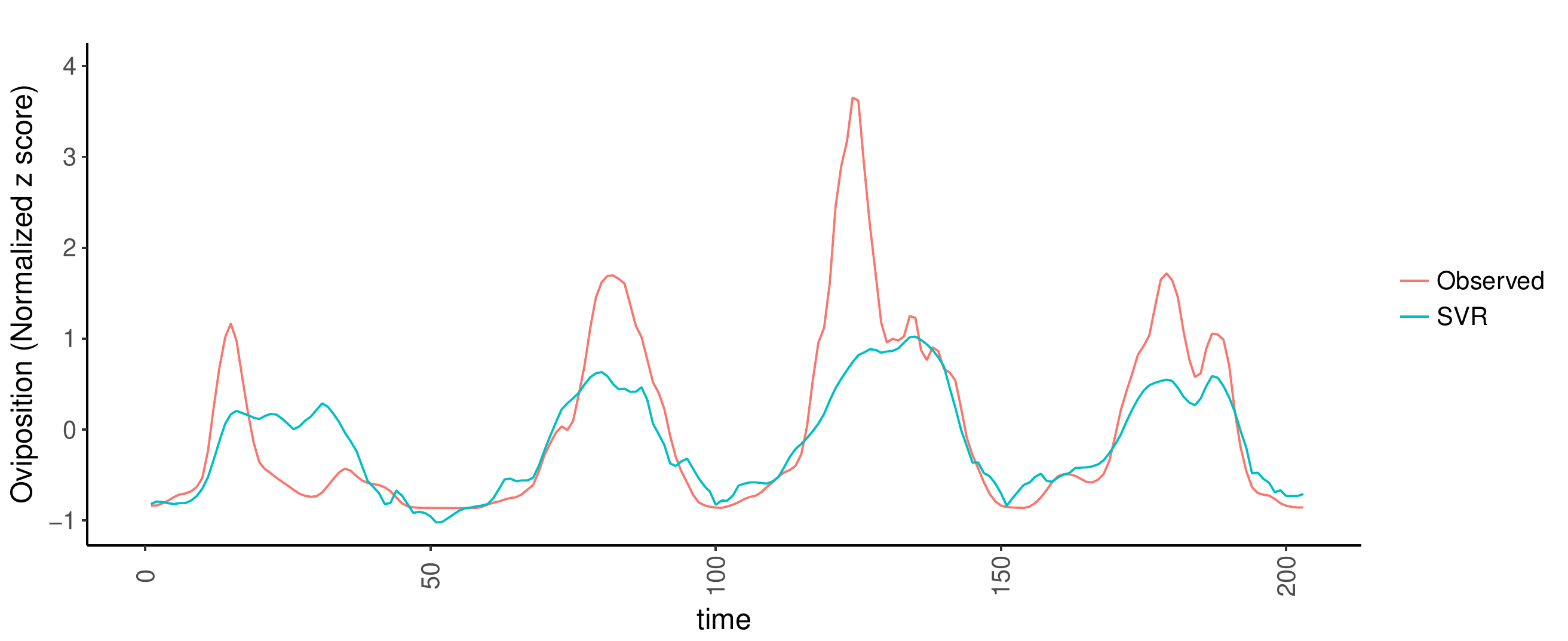}
\caption{Observed z-score and SVR regressions}\label{Fig:SVR}
\end{figure}

Figure~\ref{Fig:SVR} shows the observed data and the result of the Support Vector Regression (SVR) procedure.
The latter fails to model the peaks of the former, but produces a relatively good fit in the bulk of the data.

\begin{figure}[hbt]
\centering
\includegraphics[width=\linewidth]{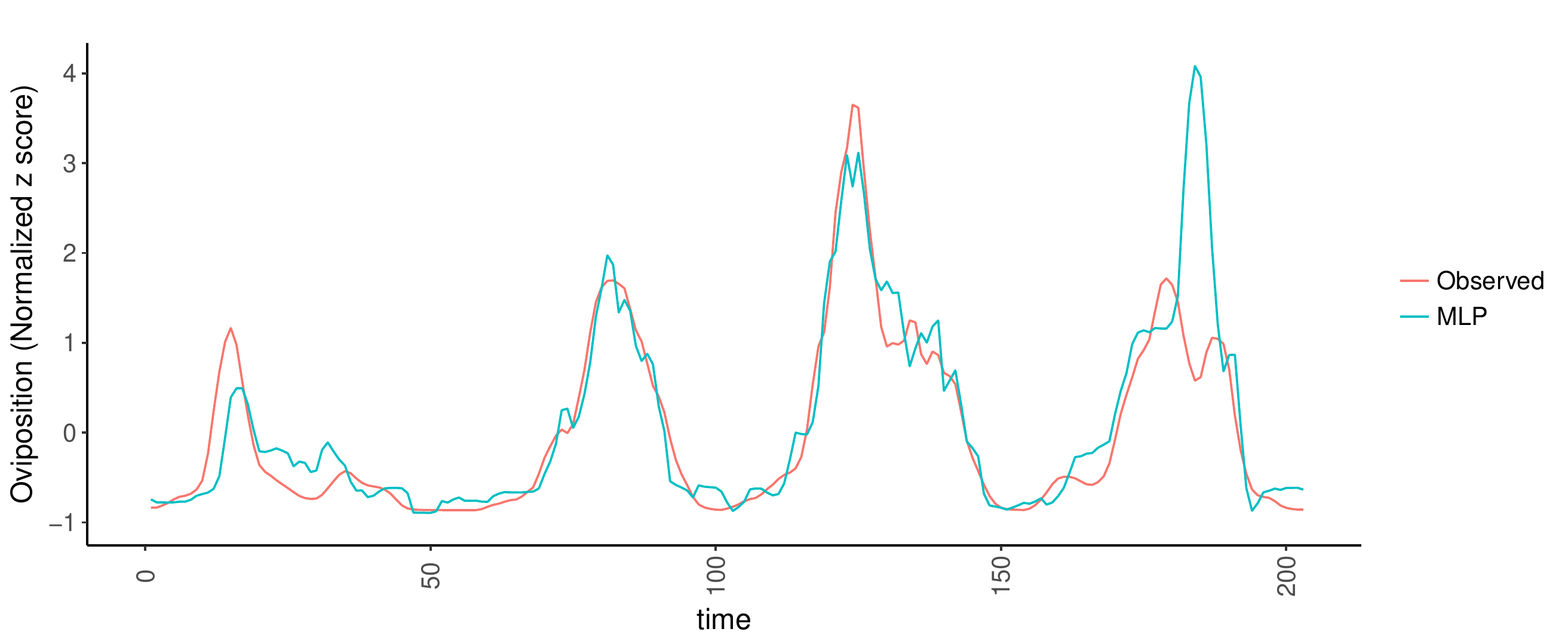}
\caption{Observed z-score MLP regression}\label{Fig:MLP}
\end{figure}

Figure~\ref{Fig:MLP} shows the results of fitting the observed data with the Multilayer Perceptron (MLP) technique.
The fit is very good, although the model overestimates the data around the twenty-fifth week of the study, and underestimates them around the last peak.

\begin{figure}[hbt]
\centering

\includegraphics[width=\linewidth]{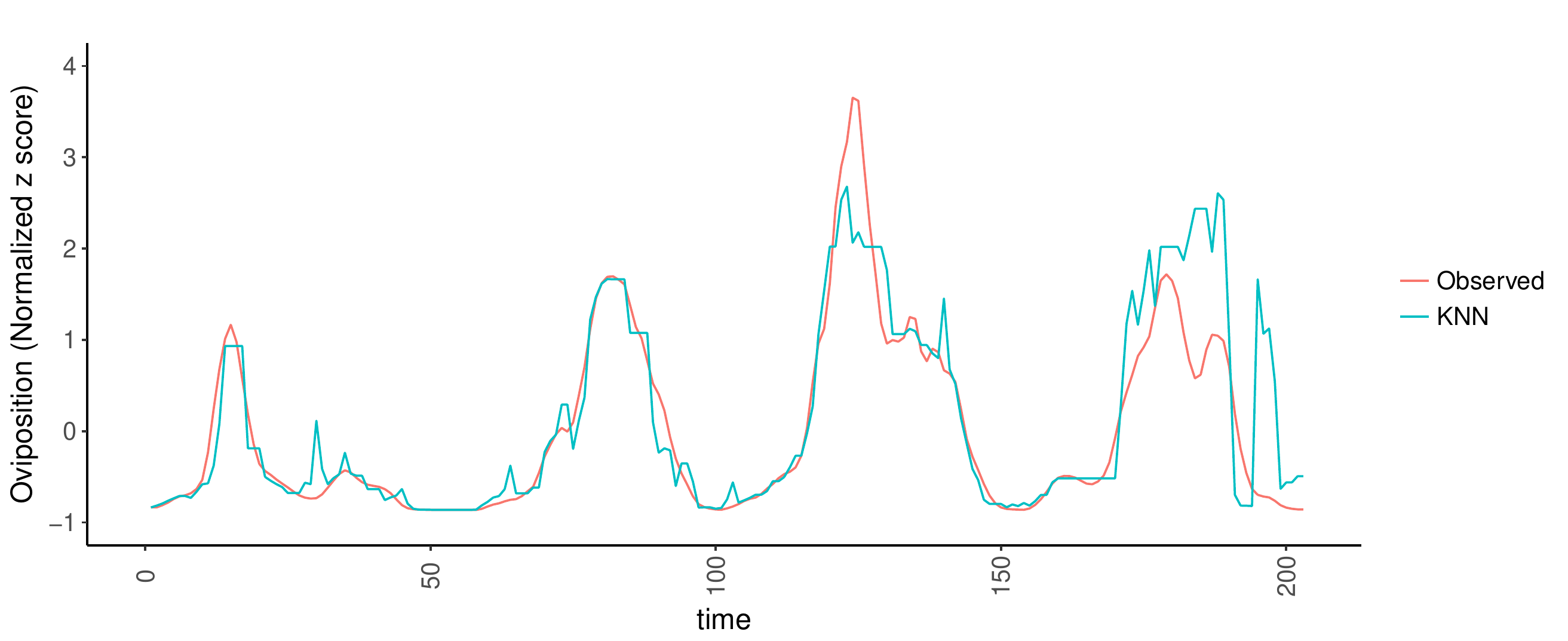}
\caption{Observed z-score and KNN regression}\label{Fig:KNN}
\end{figure}

Figure~\ref{Fig:KNN} shows the results produced by the KNN procedure.
Also, this is a very good model although it fails to follow the two largest peaks.
The first, around the 125th week is underestimated, and the second, which is close to the 180th week, is overestimated.

\begin{figure}[hbt]
\centering
\includegraphics[width=\linewidth]{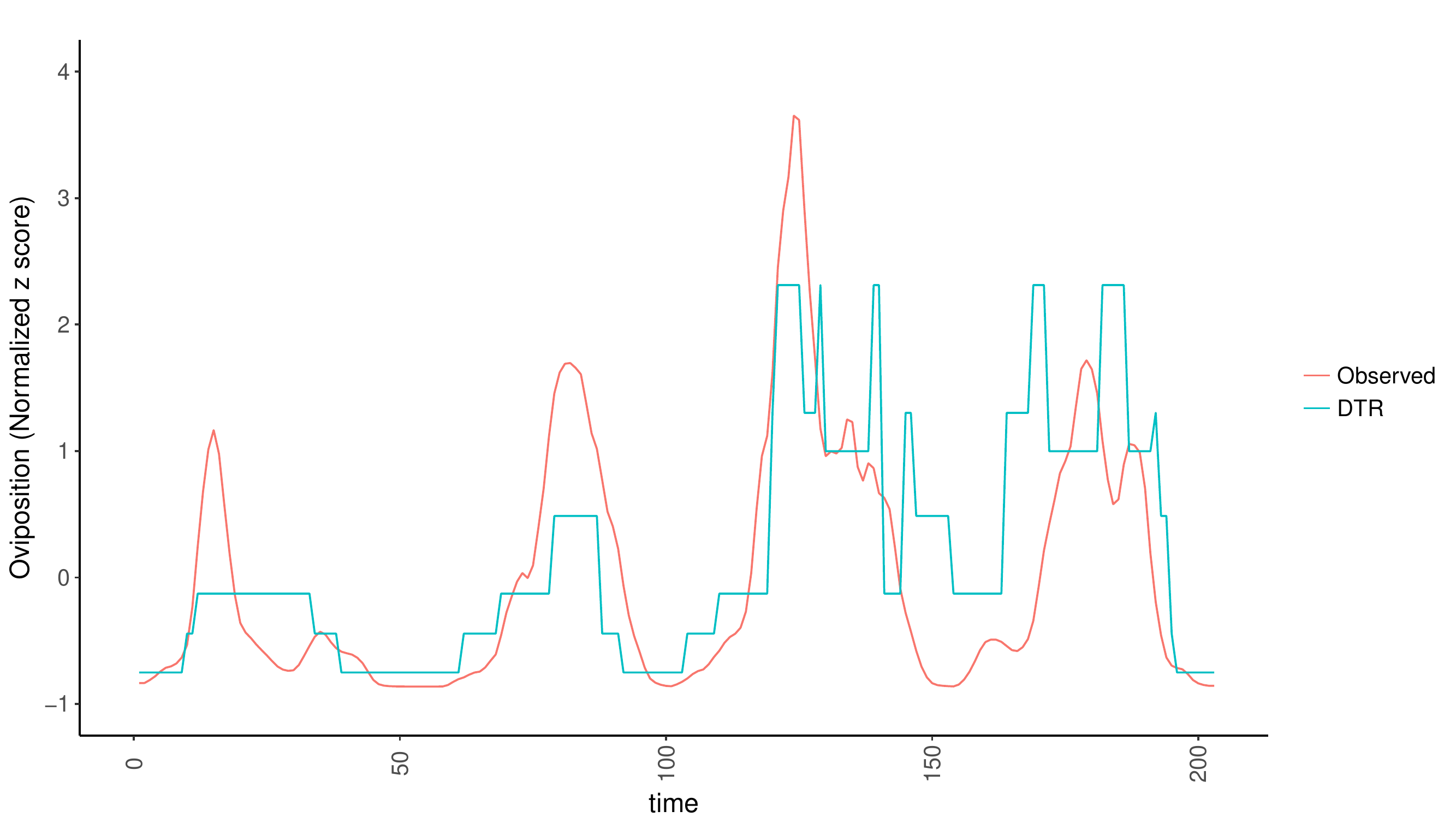}
\caption{Observed z-score and Decision Tree regression}\label{Fig:DTR}
\end{figure}

Figure~\ref{Fig:DTR} shows the result of applying the Decision Tree Regressor.
The structure of this technique produces flat outputs which, nevertheless, follow closely the observed data. 
It is important to remember that in all previous figures the last forty weeks are not used to build the models, therefore they are completely predicted.

Table~\ref{Tab:Summary} presents a summary of the observed and fitted data:
the minimum (Min) and maximum (Max) values,
the first ($q_{1/4}$) and third ($q_{3/4}$) quartiles,
the median ($q_{1/2}$) and the mean.

\begin{table}[hbt]
\centering
\caption{Summary of the observed and fitted data}\label{Tab:Summary}
\begin{tabular}{*7{r}}
\toprule
&Min	&$q_{1/4}$	&$q_{1/2}$	&Mean	&$q_{3/4}$	&Max\\ \cmidrule(lr){2-7}
Observed	&$-0.863$	&$-0.742$	&$-0.487$	&$0.000$	&$0.704$	&$3.652$\\
Linear	&$-1.641$	&$-0.716$	&$ 0.027$	&$-0.087$	&$0.462$	&$1.387$\\
Ridge	&$-1.638$	&$-0.680$	&$ 0.028$	&$-0.084$	&$0.459$	&$1.370$\\
MLP	&$-0.894$	&$-0.677$	&$-0.323$	&$0.093$	&$0.716$	&$4.084$\\
DTR	&$-0.752$	&$-0.752$	&$-0.128$	&$0.138$	&$0.998$	&$2.312$\\
KNNR	&$-0.863$	&$-0.699$	&$-0.501$	&$0.099$	&$1.033$	&$2.679$\\
SVR	&$-1.021$	&$-0.601$	&$-0.232$	&$-0.147$	&$0.309$	&$1.023$\\
\bottomrule
\end{tabular}
\end{table}

Table~\ref{Tab:Summary} reveals the following facts:
\begin{itemize}
\item Linear and Ridge regressions exaggerate the minima, as they produce values which are approximately the double of the observed ones.
\item The Multilayer Perceptron exaggerates the maximum by about \SI{10}{\percent}, while the other models underestimate it. Notice that the Support Vector Regression flattens the maximum by a factor of about \num{3.6}.
\item The mean and median of the observed data differ noticeably, suggesting that they are significantly skewed to the left.
\item The closest median value to the observed one is produced by K-Nearest Neighbors, which also leads to a very close mean value.
\end{itemize}

\begin{figure}
\centering
\includegraphics[width=.8\linewidth]{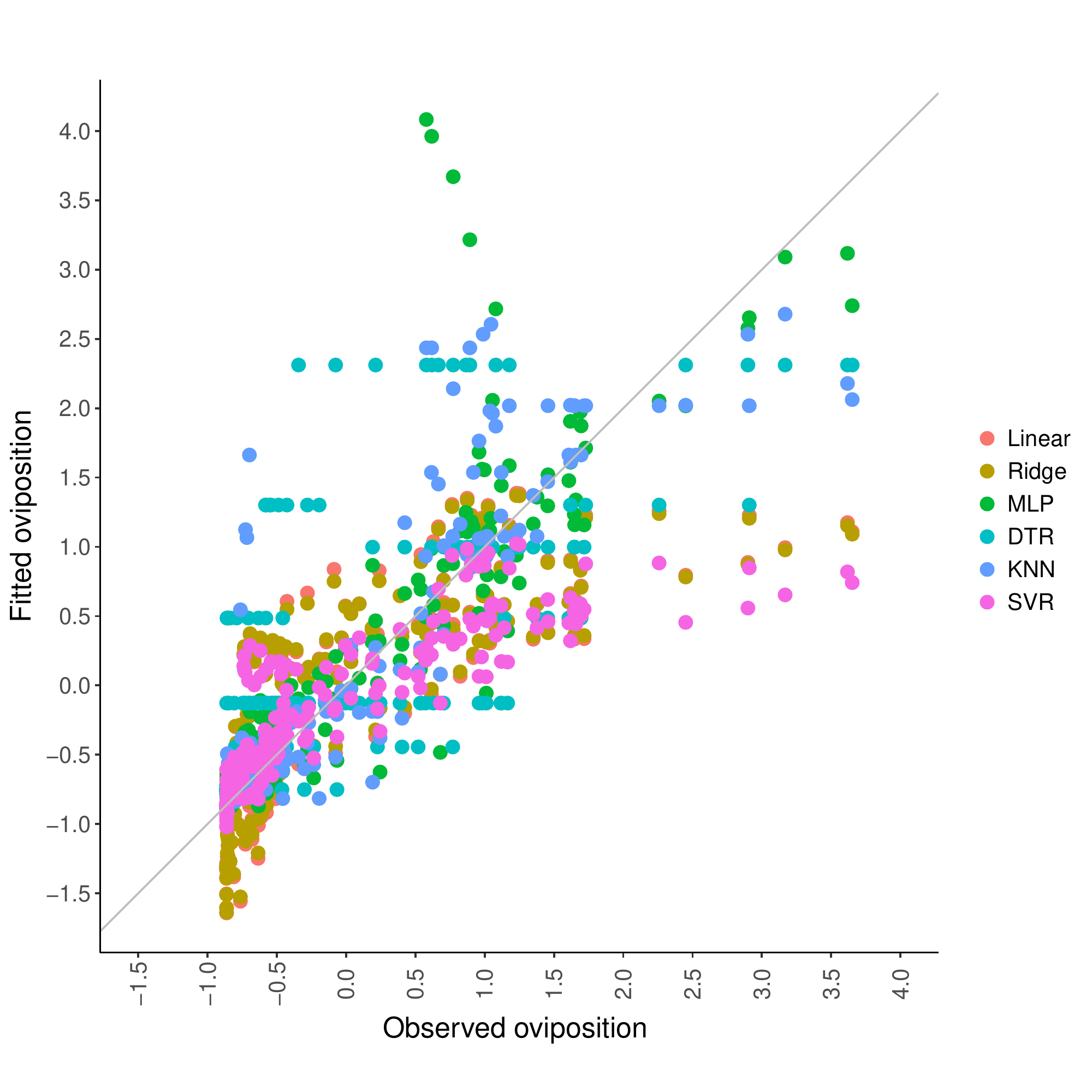}
\caption{Scatterplot of observed and predicted values}\label{Fig:ScatterPlot}
\end{figure}

Figure~\ref{Fig:ScatterPlot} shows the observed and predicted data as a scatterplot.
This figure reveals that none of the models is able to follow the largest observed values, and that the Linear, Ridge and Support Vector Regressions are the least apt for this task, while the Multilayer Perceptron is the closest one.
We also notice that this last model is the most prone to overestimating the data.
Notice that underestimation is, from the application viewpoint, more dangerous than overestimation, as the former leads to a false negative indicator that may lead to not firing preventive measures in cases when they are needed.

%%% ACF Usar los mismos colores de los boxplot, u otra tabla de colores que permita diferenciarlos mejor

\begin{figure}
\centering
\subfigure[Histograms\label{Fig:Histograms}]{\includegraphics[width=.8\linewidth]{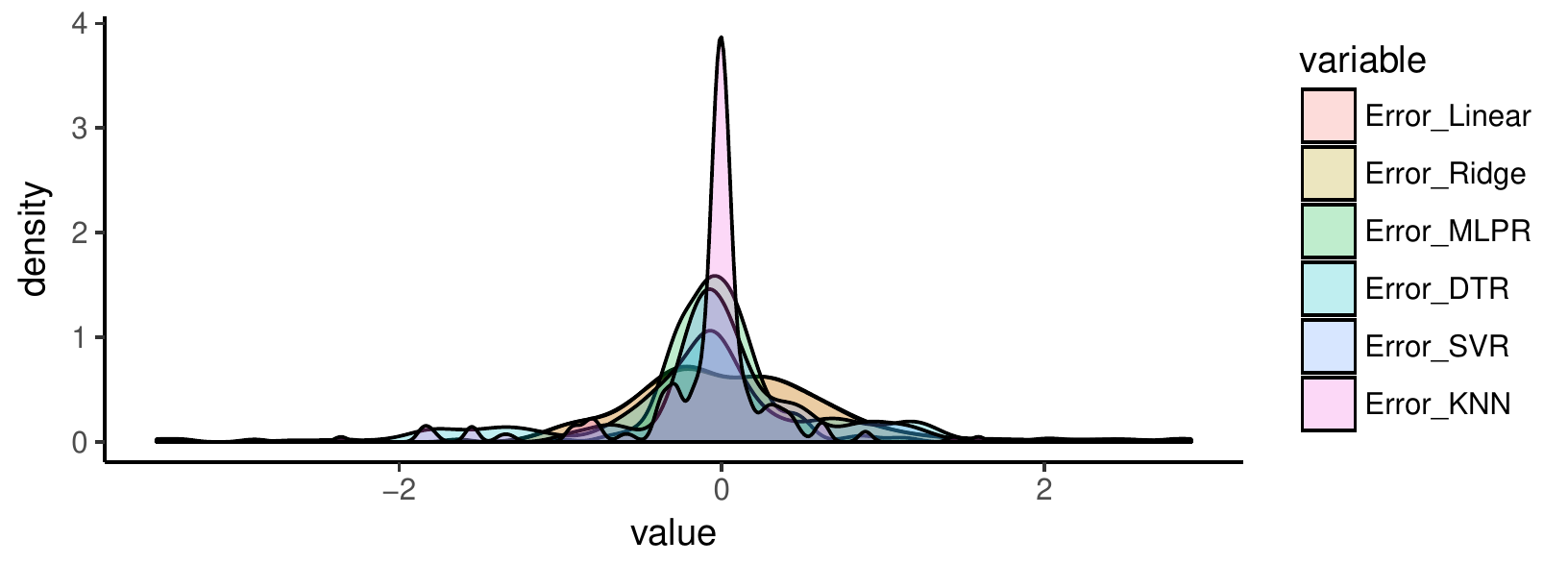}}
\subfigure[Boxplots\label{Fig:Boxplots}]{\includegraphics[width=.8\linewidth]{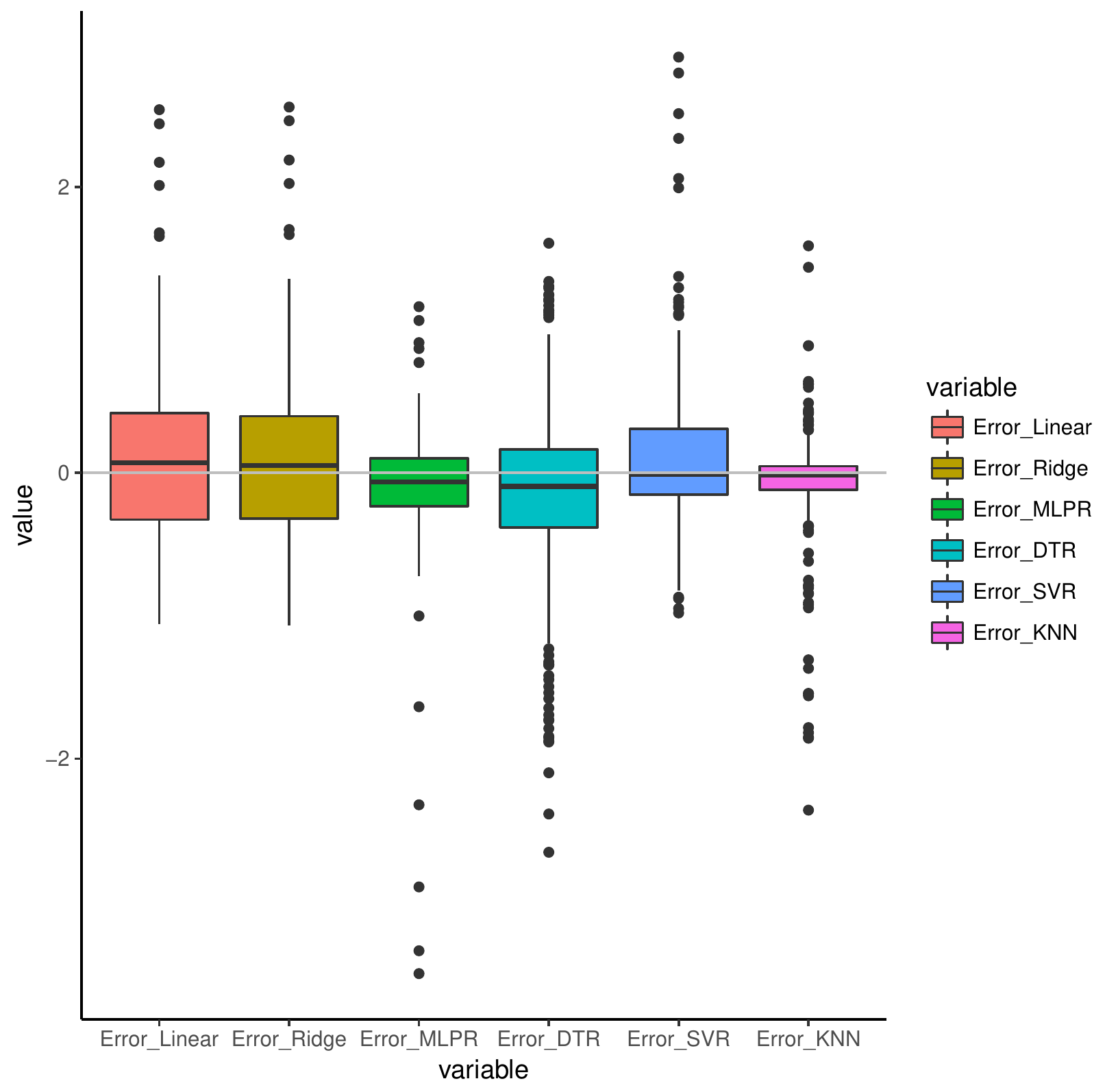}}
\caption{Residuals}\label{Fig:Residuals}
\end{figure}

In the following we analyze the residuals.
Figures~\ref{Fig:Histograms} and~\ref{Fig:Boxplots} show, respectively, the histograms and boxplots of the errors produced by each model.
The errors produced by KNN are the most concentrated around zero, followed by MLP.
The two errors most spread are due to the linear regressions.
This is an indication that the models obtained using simple linear techniques are the worst among the ones considered here.
%%% ACF Hay algo raro. En la tabla consta que la mediana más lejana del cero es KNNR, pero en el gráfico de boxplots parece ser DTR. Va así, pero hay que revisar.

Table~\ref{Tab:Quality} presents quality measures of the models here considered:
Pearson correlation coefficients between the observed and fitted values, using the complete data set (Corr11) and the \SI{20}{\percent} (CorrL20) left for validation; and
the Mean Square Error of the complete data set (MSE) and of the validation data (MSEL20).
Following \cite{Cramer2017}, we also include the mean Score obtained from the cross validation and its standard deviation.

\begin{table}[hbt]
\centering
\caption{Measures of quality of the models}\label{Tab:Quality}
\begin{tabular}{*7{r}}
\toprule
& Corr11
& MSE
& Mean Score
& SD of Score
& CorrL20
& MSEL20 \\ \midrule
Linear
&$0.774$
&$0.624$
&$1.108$%&\boldmath$1.108$  %CARO: cambio
&$0.278$
&$0.890$
&$0.580$\\
Ridge
&$0.775$
&$0.621$
&$1.072$
&$0.277$
&$0.896$
&$0.566$ \\
SVR
&$0.837$
&$0.613$
&\boldmath$0.834$   %CARO: cambio
&$0.490$
&\boldmath$0.967$
&\boldmath$0.464$ \\
MLP
&$0.875$
&$0.528$
&$1.086$
&$0.288$
&$0.727$
&$1.023$ \\
KNN
&\boldmath$0.888$
&\boldmath$0.494$
&$0.981$
&$0.362$
&$0.797$
&$0.936$\\
DTR
&$0.679$
&$0.768$
&$1.148$
&$0.544$
&$0.532$
&$1.131$ \\
\bottomrule
\end{tabular}
\end{table}

\section{Discussion}

Taking into account all the goodness-of-fit parameters included in Tables~\ref{Tab:Summary} and~\ref{Tab:Quality}, the figures \ref{Fig:Boxplots} and \ref{Fig:Residuals} and the analysis of errors, we may consider that KNN stems as the best method for this problem. 
It has a correlation near to \SI{90}{\percent}, considerably greater than \SI{75}{\percent}, the typical values obtained with linear approaches. 

%NOTA Caro: cambio Linear regression por Super Vector Regression y agrego cita de cramer 2017 para justificar.

The Mean Score value would lead to choose Support Vector Regressor %Linear Regression 
as the best technique \cite{Cramer2017}.
It is noteworthy that the standard deviation of this measure of quality is so high that is it unlikely that it is able to render a good choice by itself.
For this reason, we follow a holistic approach in the forthcoming Conclusions.

%In this kind of modeling, basically we have three large issues to consider. The response variable that we want to represent, in this case number of eggs. It is clear that, based in ovitraps measurements, we could consider others alternatives like proportion positive traps for example. The second one is the independent variables that we choose like predictors (Ndvi, ndwi, etc). And finally the method that to use to generate the model (Linear, Generalized linear, logistic, neural networks etc.). As we said this paper dont focus on the two first issues and then those are based in previous results/publications. Of course these are matters of discussion and could be improved. Here, in this contribution, we prefer leave these points untouched and focus our attention on the possibilities of the different machine learning tools to model those variables. In that sense it should be clear that here we are not closing the general problem of modeling vector populations. 

An interesting point that appears in the results of all the models here presented, is that that models 
fit well the main pattern but not necessarily the large peaks. 
One hypothesis is that the vector population may disengage the macro-environmental/climatic variables when conditions are optimal and, again, be restricted when the environmental conditions get poorer. 
In fact, it would be clear that, we can not hope to fit exactly this urban vector population only based in large scale macro-environmental variables.

%It is clear that this kind of machine learning models or data based models have a fundamental limitation related with the universe of data used for build the relations between the variables. So, as all statistical model, those will be valid only while the conditions that gave rise to the data used to build the model continue to be fulfilled.

\section{Conclusions}

Dengue, Chikungunya and Zika are viral diseases for which there is no vaccine. Therefore, the most effective control comes from preventing the spread of \textit{Ae.\ \ae gypti} (Linnaeus) and, thus, knowing about its population dynamics is of paramount importance. 
This work provides a framework for forecasting oviposition using remotely sensed variables solely, and freely available Machine Learning tools. 
Such tools are improving the Argentinean operational risk system \cite{Porcasi2012}.

We used operationally available satellite derived environmental variables (temperature, humidity and precipitation) to build  temporal models able to predict the oviposition activity outside the houses.
In this way, our perspective completely operative, means generate a procedure to estimate the vector activity and then eventually become independent of field measurements,(not predict the future), considering that to measure oviposition in 50 houses all the weeks all the time (as we use to generate the model) have a very large cost .

%Also different time lags variables were included due to their importance for forecast. 

%The dataset spans over \num{4} years and, with this, it allows building a more general model. These variables involved can then be incorporated into an operational risk stratification framework.

This study improved previous epidemiology studies, which consider statistical models with linear relationships \cite{Estallo2008, Estallo2012,Estallo2016}. 
Such improvement is obtained by the use of Machine Learning tools that impose the user no significant additional effort. 

The proposal showed that out-of-the-shelf FLOSS tools are capable of dealing with the complex relationships among variables providing, thus, an almost effortless and free way of handling with this relevant problem.
This interdisciplinary approach provides new tools for practitioners.

This work is an example of how the use of automatic algorithm configuration tools like \verb|iRace| can reduce the complexity of parameter tuning and provide a frame of reference for model selection. 
Additionally, we show the importance of training with Cross Validation, a commodity in image classification but seldom used by Remote Sensing operative users. 
Cross validation was used in order to decrease the dependency of the evaluation results on a particular selection of training set and validation set pair. In particular, a time series split cross validation procedure was used to evaluate the model. 
All the models here discussed can be run with a Python script freely available at \url{https://github.com/JuanScaFranTru/mosquitomodels}.
%%% ACF Hay que dejar el script disponible antes de mandar el artículo 

%Typically the remote sensing applications developed by users (out the academic walls) are based exclusively on the use of software package free or proprietary like Envi, qgis, grass, Erdas and on classical statistical methodologies. Here we trait to show that it is possible to use open source, and easy to use, programing framework like offered by Python and R specific libs (scikit learn, numpy, scipy, iRace, opencv, GDal ), enabling include in the studies of the RS users community moderns and powerful tools. 

We found that K-Nearest Neighbour Regression (KNNR), MLP and SVM  improve predictive models of vector population based on satellite derived environmental variables. 
The performance of these algorithms could be improved substantially using a larger dataset. 
Although the used period is large in comparison with similar works on vector population, at the same time the used dataset is very small from the machine learning point of view.

Finally this work presents several improvements regarding previous works \cite{Estallo2012,Estallo2016,German2017,Estallo2014}, in terms of temporal data length,  
the use a more complete accessible operatively set of remotely sensed variables, and
mostly with respect to the use of ML learning modeling.

%%%%% Version original
%Taking into account the importance of vector populations control actions regarding the prevention, then to generate knowledge about the temporal  dynamic of \textit{Aedes aegypti}(Linnaeus) on time is fundamental for planning control activities and resources.  
%%% Version alternativa
As the control of vector populations is a very important task in prevention of diseases, the knowledge of temporal dynamics of \textit{Aedes aegypti} (Linnaeus) plays a fundamental role in planning strategies and resources management.
%%%%
The result presented in this work, with models producing a correlation about \SI{90}{\percent} with the actual oviposition temporal series, show the usefulness of a procedure based on satellite derived environmental variables and easily accessible modern machine learning techniques. 
Applying the same procedure to other cities is likely to lead to an improved national operational risk system \citep{Porcasi2012}. 
In addition, the kind of data and tools presented (all freely accessible) allow the replication of the methodology, if not the same model, in  other regions. 

%\cite{IsDenoisingDead}

\section{References}

%\biboptions{comma,square,authoryear}  % NOTA CARO: Por algun motivo no funcionan estas opciones
%\bibliographystyle{model2-names.bst}
\bibliographystyle{elsarticle-num}     % NOTA CARO: pongo este formato que funciona, cambiar si lo desean.
\bibliography{sample}

\begin{thebibliography}{10}
\expandafter\ifx\csname url\endcsname\relax
  \def\url#1{\texttt{#1}}\fi
\expandafter\ifx\csname urlprefix\endcsname\relax\def\urlprefix{URL }\fi
\expandafter\ifx\csname href\endcsname\relax
  \def\href#1#2{#2} \def\path#1{#1}\fi

\bibitem{Lary2009}
D.~J. Lary, L.~A. Remer, D.~MacNeill, B.~Roscoe, S.~Paradise, Machine learning
  and bias correction of {MODIS} aerosol optical depth, IEEE Geoscience and
  Remote Sensing Letters 6~(4) (2009) 694--698.
\newblock \href {http://dx.doi.org/10.1109/LGRS.2009.2023605}
  {\path{doi:10.1109/LGRS.2009.2023605}}.

\bibitem{Brown2008}
M.~E. Brown, D.~J. Lary, A.~Vrieling, D.~Stathakis, H.~Mussa, Neural networks
  as a tool for constructing continuous {NDVI} time series from {AVHRR} and
  {MODIS}, International Journal of Remote Sensing 29~(24) (2008) 7141--7158.
\newblock \href {http://dx.doi.org/10.1080/01431160802238435}
  {\path{doi:10.1080/01431160802238435}}.

\bibitem{Azamathulla2012}
H.~M. Azamathulla, A.~{Ab Ghani}, S.~Y. Fei, {ANFIS}-based approach for
  predicting sediment transport in clean sewer, Applied Soft Computing Journal
  12~(3) (2012) 1227--1230.

\bibitem{Zahabiyoun2013}
B.~Zahabiyoun, M.~R. Goodarzi, A.~R.~M. Bavani, H.~M. Azamathulla, Assessment
  of climate change impact on the {G}haresou river basin using {SWAT}
  hydrological model, CLEAN – Soil, Air, Water 41~(6) (2013) 601--609.
\newblock \href {http://dx.doi.org/10.1002/clen.201100652}
  {\path{doi:10.1002/clen.201100652}}.

\bibitem{Madadi2015}
M.~R. Madadi, H.~M. Azamathulla, M.~Yakhkeshi, Application of {G}oogle {E}arth
  to investigate the change of flood inundation area due to flood detention
  dam, Earth Science Informatics 8~(3) (2015) 627--638.

\bibitem{Yi1996}
J.~Yi, V.~R. Prybutok, A neural network model forecasting for prediction of
  daily maximum ozone concentration in an industrialized urban area,
  Environmental Pollution 92~(3) (1996) 349 -- 357.
\newblock \href {http://dx.doi.org/10.1016/0269-7491(95)00078-X}
  {\path{doi:10.1016/0269-7491(95)00078-X}}.

\bibitem{Lary2016}
D.~J. Lary, A.~H. Alavi, A.~H. Gandomi, A.~L. Walker, Machine learning in
  geosciences and remote sensing, Geoscience Frontiers 7~(1) (2016) 3 -- 10,
  special Issue: Progress of Machine Learning in Geosciences.
\newblock \href {http://dx.doi.org/10.1016/j.gsf.2015.07.003}
  {\path{doi:10.1016/j.gsf.2015.07.003}}.

\bibitem{Penia2014}
J.~Pe{\~n}a-Barrag{\'a}n, P.~A. Guti{\'e}rrez, C.~Herv{\'a}s-Mart{\'i}nez,
  J.~Six, R.~E. Plant, F.~L{\'o}pez-Granados, {O}bject-{B}ased {I}mage
  {C}lassification of {S}ummer {C}rops with {M}achine {L}earning {M}ethods,
  Remote Sensing 6~(6) (2014) 5019--5041.
\newblock \href {http://dx.doi.org/10.3390/rs6065019}
  {\path{doi:10.3390/rs6065019}}.

\bibitem{Atkinson1997}
P.~M. Atkinson, A.~R.~L. Tatnall, Introduction neural networks in remote
  sensing, International Journal of Remote Sensing 18~(4) (1997) 699--709.

\bibitem{Zhang2005}
G.~Zhang, M.~Qi, Neural network forecasting for seasonal and trend time series,
  European Journal of Operational Research 160~(2) (2005) 501 -- 514.
\newblock \href {http://dx.doi.org/10.1016/j.ejor.2003.08.037}
  {\path{doi:10.1016/j.ejor.2003.08.037}}.

\bibitem{Foody2004}
G.~M. Foody, Supervised image classification by {MLP} and {RBF} neural networks
  with and without an exhaustively defined set of classes, International
  Journal of Remote Sensing 25~(15) (2004) 3091--3104.
\newblock \href {http://dx.doi.org/10.1080/01431160310001648019}
  {\path{doi:10.1080/01431160310001648019}}.

\bibitem{Bose2017}
P.~Bose, N.~K. Kasabov, L.~Bruzzone, R.~N. Hartono, Spiking neural networks for
  crop yield estimation based on spatiotemporal analysis of image time series,
  IEEE Transactions on Geoscience and Remote Sensing 54~(11) (2016) 6563--6573.
\newblock \href {http://dx.doi.org/10.1109/TGRS.2016.2586602}
  {\path{doi:10.1109/TGRS.2016.2586602}}.

\bibitem{Wang2016}
D.~Wang, Y.~Li, B.~Gao, Neural network technology and semi-analytical approach
  combined model for remote sensing chlorophyll-a concentration, in: 2016 IEEE
  International Geoscience and Remote Sensing Symposium (IGARSS), 2016, pp.
  5852--5855.
\newblock \href {http://dx.doi.org/10.1109/IGARSS.2016.7730529}
  {\path{doi:10.1109/IGARSS.2016.7730529}}.

\bibitem{JafariGoldarag2016}
Y.~Jafari~Goldarag, A.~Mohammadzadeh, A.~S. Ardakani, Fire risk assessment
  using neural network and logistic regression, Journal of the Indian Society
  of Remote Sensing 44~(6) (2016) 885--894.
\newblock \href {http://dx.doi.org/10.1007/s12524-016-0557-6}
  {\path{doi:10.1007/s12524-016-0557-6}}.

\bibitem{Powell2013}
J.~Powell, W.~Tabachnick, History of domestication and spread of aedes
  aegypti--a review, Memórias do Instituto Oswaldo Cruz 108 (2013) 11--17.
\newblock \href {http://dx.doi.org/10.1590/0074-0276130395}
  {\path{doi:10.1590/0074-0276130395}}.

\bibitem{Moncayo2004}
A.~Moncayo, Z.~Fernandez, D.~Ortiz, M.~Diallo, A.~Sall, S.~Hartman, C.~Davis,
  L.~Coffey, C.~Mathiot, R.~Tesh, S.~Weaver, Dengue emergence and adaptation to
  peridomestic mosquitoes, Emerging Infectious Diseases 10~(10).

\bibitem{WHO2015}
WHO, \href{http://www.who.int/mediacentre/factsheets/fs117/en/}{Dengue and
  severe dengue, {F}act sheet no 117, {U}p-dated {M}ay 2015} (2015).
\newline\urlprefix\url{http://www.who.int/mediacentre/factsheets/fs117/en/}

\bibitem{who2015b}
WHO, \href{http://www.who.int/mediacentre/factsheets/fs327/en/}{Chikungunya,
  fact sheet no 327, updated may 2015} (2015).
\newline\urlprefix\url{http://www.who.int/mediacentre/factsheets/fs327/en/}

\bibitem{who2016}
WHO, \href{http://www.who.int/mediacentre/factsheets/zika/en/}{Zika virus, fact
  sheet january 2016} (2015).
\newline\urlprefix\url{http://www.who.int/mediacentre/factsheets/zika/en/}

\bibitem{Ritchie1084}
S.~Ritchie, The production of {A}edes aegypti by a weekly ovitraps survey,
  Mosquito News 44~(1) (1984) 77--79.

\bibitem{Ostefeld2005}
R.~S. Ostfeld, G.~E. Glass, F.~Keesing, Spatial epidemiology: an emerging (or
  re-emerging) discipline, Trends in Ecology \& Evolution 20~(6) (2005) 328 --
  336.
\newblock \href {http://dx.doi.org/10.1016/j.tree.2005.03.009}
  {\path{doi:10.1016/j.tree.2005.03.009}}.

\bibitem{Pavlovsky1966}
E.~Pavlovky, Natural nidality of transmissible diseases with special reference
  to the landscape epidemiology of zooanthroponoses, University of Illinois
  Press, Urbana.

\bibitem{Hay2000}
S.~Hay, An overview of remote sensing and geodesy for epidemiology and public
  health application, in: Remote Sensing and Geographical Information Systems
  in Epidemiology, Vol.~47 of Advances in Parasitology, Academic Press, 2000,
  pp. 1--35.
\newblock \href {http://dx.doi.org/10.1016/S0065-308X(00)47005-3}
  {\path{doi:10.1016/S0065-308X(00)47005-3}}.

\bibitem{Parra2010}
G.~Parra-Enao, Sistemas de informacion geografica y sensores remotos.
  aplicaciones en enfermedades transmitidas por vectores, Rev CES Med 24~(2)
  (2010) 75--90.

\bibitem{Douglas2010}
D.~Fuller, A.~Troyo, O.~Calderon-Arguedas, J.~C. Beier, Dengue vector (aedes
  aegypti) larval habitats in an urban environment of costa rica analysed with
  aster and quickbird imagery, International Journal of Remote Sensing 31~(1)
  (2010) 3--11.
\newblock \href {http://dx.doi.org/10.1080/01431160902865756}
  {\path{doi:10.1080/01431160902865756}}.

\bibitem{moreno2014}
M.~Moreno-Madrinán, W.~Crosson, Correlating remote sensing data with the
  abundance of pupae of the dengue virus mosquito vector, aedes aegypti, in
  central mexico, ISPRS International Journal of Geo-Information 3~(2) (2014)
  732--749.

\bibitem{arboleda2012}
S.~Arboleda, N.~Jaramillo-O, A.~Peterson, Spatial and temporal dynamics of
  aedes aegypti larval sites in bello, colombia, Journal of Vector Ecology
  37~(1) (2012) 37--48.
\newblock \href {http://dx.doi.org/10.1111/j.1948-7134.2012.00198.x}
  {\path{doi:10.1111/j.1948-7134.2012.00198.x}}.

\bibitem{Rotela2007}
C.~Rotela, F.~Fouque, M.~Lamfri, P.~Sabatier, V.~Introini, M.~Zaidenberg,
  C.~Scavuzzo, Space-time analysis of the dengue spreading dynamics in the 2004
  {T}artagal outbreak, northern {A}rgentina, Acta Tropica 103~(1) (2007) 1--13.

\bibitem{Estallo2011}
E.~L. Estallo, F.~F. {Ludue{\~n}a-Almeida}, A.~M. Visintin, C.~M. Scavuzzo,
  M.~V. Introini, M.~Zaidenberg, W.~R. Almir\'on, Prevention of dengue
  outbreaks through aedes aegypti oviposition activity forecasting method,
  Vector-Borne and Zoonotic Diseases 11~(5) (2011) 543--549.

\bibitem{Espinosa2016b}
M.~O. Espinosa, D.~Weinberg, C.~H. Rotela, F.~Polop, M.~Abril, C.~M. Scavuzzo,
  Temporal dynamics and spatial patterns of {A}edes aegypti breeding sites, in
  the context of a dengue control program in {T}artagal ({S}alta province,
  {A}rgentina), PLoS Neglected Tropical Diseases 10~(5).

\bibitem{Porcasi2012}
X.~Porcasi, C.~H. Rotela, M.~V. Introini, N.~Frutos, S.~Lanfri, G.~Peralta,
  E.~A. De~Elia, M.~A. Lanfri, C.~M. Scavuzzo, An operative dengue risk
  stratification system in {A}rgentina based on geospatial technology,
  Geospatial Health 6~(3 SUPPL.) (2012) S31--S42.

\bibitem{Herbreteau2007}
V.~Herbreteau, G.~Salem, M.~Souris, J.-P. Hugot, J.-P. Gonzalez, Thirty years
  of use and improvement of remote sensing, applied to epidemiology: From early
  promises to lasting frustration, Health and Place 13~(2) (2007) 400--403.
\newblock \href {http://dx.doi.org/10.1016/j.healthplace.2006.03.003}
  {\path{doi:10.1016/j.healthplace.2006.03.003}}.

\bibitem{Kalluri2007}
S.~Kalluri, P.~Gilruth, D.~Rogers, M.~Szczur, Surveillance of arthropod
  vector-borne infectious diseases using remote sensing techniques: A review,
  PLOS Pathogens 3~(10) (2007) 1--11.
\newblock \href {http://dx.doi.org/10.1371/journal.ppat.0030116}
  {\path{doi:10.1371/journal.ppat.0030116}}.

\bibitem{Buczak2012}
A.~Buczak, P.~Koshute, S.~Babin, B.~Feighner, S.~Lewis, A data-driven
  epidemiological prediction method for dengue outbreaks using local and remote
  sensing data, BMC Medical Informatics and Decision Making 12~(1) (2012) 124.
\newblock \href {http://dx.doi.org/10.1186/1472-6947-12-124}
  {\path{doi:10.1186/1472-6947-12-124}}.

\bibitem{Bowman2016}
L.~Bowman, G.~Tejeda, G.~Coelho, L.~Sulaiman, B.~Gill, P.~McCall, P.~Olliaro,
  S.~Ranzinger, L.~Quang, R.~Ramm, A.~Kroeger, M.~Petzold, Alarm variables for
  dengue outbreaks: A multi-centre study in asia and latin america, PLoS ONE
  11~(6).
\newblock \href {http://dx.doi.org/10.1371/journal.pone.0157971}
  {\path{doi:10.1371/journal.pone.0157971}}.

\bibitem{Estallo2012}
E.~L. Estallo, F.~F. {Ludue\~na}-Almeida, A.~M. Visintin, C.~M. Scavuzzo, M.~A.
  Lamfri, M.~V. Introini, M.~Zaidenberg, W.~R. Almir\'on, Effectiveness of
  normalized difference water index in modelling {A}edes aegypti house index,
  International Journal of Remote Sensing 33~(13) (2012) 4254--4265.
\newblock \href {http://dx.doi.org/10.1080/01431161.2011.640962}
  {\path{doi:10.1080/01431161.2011.640962}}.

\bibitem{Estallo2016}
E.~L. Estallo, E.~M. Benitez, M.~A. Lanfri, C.~M. Scavuzzo, W.~R. Almir\'on,
  {MODIS} environmental data to assess {C}hikungunya, {D}engue, and {Z}ika
  diseases through {A}edes ({S}tegomia) aegypti oviposition activity
  estimation, IEEE Journal of Selected Topics in Applied Earth Observations and
  Remote Sensing 9~(12) (2016) 5461--5466.
\newblock \href {http://dx.doi.org/10.1109/JSTARS.2016.2604577}
  {\path{doi:10.1109/JSTARS.2016.2604577}}.

\bibitem{German2017}
M.~O. Espinosa, E.~Alvarez Di~Fino, M.~Abril, M.~A. Lanfri, M.~V. Periago,
  C.~M. Scavuzzo, Operational satellite based temporal modeling of aedes
  population, Sent to Geospatial Health 2017.

\bibitem{Rotela2017}
C.~Rotela, L.~Lopez, M.~Fr\'ias~C\'espedes, A.~Lighezzolo, X.~Porcasi,
  M.~Lanfri, C.~Scavuzzo, D.~Gorla, Analytical report of the 2016 dengue
  outbreak in {C}\'ordoba city, {A}rgentina, Geospatial Health 12.

\bibitem{Gomes1998}
A.~C. Gomes, Medidas dos niveis de infestacao urbana para aedes (stegomyia)
  aegypti e aedes (stegomyia) albopictus em {P}rograma de {V}igilancia
  {E}ntomol\'ogica, Informe Epidemiol\'ogico do Sus 7 (1998) 49 -- 57.
\newblock \href {http://dx.doi.org/10.5123/S0104-16731998000300006}
  {\path{doi:10.5123/S0104-16731998000300006}}.

\bibitem{Hay1997}
S.~I. Hay, M.~J. Packer, D.~J. Rogers, Review article the impact of remote
  sensing on the study and control of invertebrate intermediate hosts and
  vectors for disease, International Journal of Remote Sensing 18~(14) (1997)
  2899--2930.

\bibitem{Gao1996}
B.~Gao, {NDWI}--{A} normalized difference water index for remote sensing of
  vegetation liquid water from space, Remote Sensing of Environment 58~(3)
  (1996) 257 -- 266.
\newblock \href {http://dx.doi.org/10.1016/S0034-4257(96)00067-3}
  {\path{doi:10.1016/S0034-4257(96)00067-3}}.

\bibitem{Peres2004}
L.~F. Peres, C.~C. DaCamara, Land surface temperature and emissivity estimation
  based on the two-temperature method: sensitivity analysis using simulated
  {MSG/SEVIRI} data, Remote Sensing of Environment 91~(3) (2004) 377 -- 389.
\newblock \href {http://dx.doi.org/10.1016/j.rse.2004.03.011}
  {\path{doi:10.1016/j.rse.2004.03.011}}.

\bibitem{Wan1999}
Z.~Wan, {MODIS} land-surface temperature algorithm theoretical basis document
  ({LST ATBD}), Institute of Computational Earth System Science.

\bibitem{Wan2004}
Z.~Wan, Y.~Zhang, Q.~Zhang, Z.-L. Li, Quality assessment and validation of the
  {MODIS} global land surface temperature, International Journal of Remote
  Sensing 25~(1) (2004) 261--274.
\newblock \href {http://dx.doi.org/10.1080/0143116031000116417}
  {\path{doi:10.1080/0143116031000116417}}.

\bibitem{Kummerow1998}
C.~Kummerow, W.~Barnes, T.~Kozu, J.~Shiue, J.~Simpson, The {T}ropical
  {R}ainfall {M}easuring {M}ission ({TRMM}) sensor package, Journal of
  Atmospheric and Oceanic Technology 15~(3) (1998) 809--817.
\newblock \href
  {http://dx.doi.org/10.1175/1520-0426(1998)015<0809:TTRMMT>2.0.CO;2}
  {\path{doi:10.1175/1520-0426(1998)015<0809:TTRMMT>2.0.CO;2}}.

\bibitem{Estallo2008}
E.~Estallo, M.~Lamfri, C.~Scavuzzo, F.~Almeida, M.~Introini, M.~Zaidenberg,
  W.~Almiron, Models for predicting aedes aegypti larval indices based on
  satellite images and climatic variables, Journal of the American Mosquito
  Control Association 24~(3) (2008) 368--376.
\newblock \href {http://dx.doi.org/10.2987/5705.1} {\path{doi:10.2987/5705.1}}.

\bibitem{Estallo2014}
E.~L. Estallo, A.~E. Carbajo, M.~G. Grech, M.~Fr\'ias-C\'espedes, L.~L\'opez,
  M.~A. Lanfri, F.~F. Ludue\~na Almeida, W.~R. Almir\'on, Spatio-temporal
  dynamics of dengue 2009 outbreak in {C}\'ordoba city, {A}rgentina, Acta
  Tropica 136~(1) (2014) 129--136.

\bibitem{Cramer2017}
S.~Cramer, M.~Kampouridis, A.~A. Freitas, A.~K. Alexandridis, An extensive
  evaluation of seven machine learning methods for rainfall prediction in
  weather derivatives, Expert Systems with Applications 85 (2017) 169--181.

\bibitem{Picard1984}
R.~R. Picard, R.~D. Cook, Cross-validation of regression models, Journal of the
  American Statistical Association 79~(387) (1984) 575--583.
\newblock \href {http://dx.doi.org/10.1080/01621459.1984.10478083}
  {\path{doi:10.1080/01621459.1984.10478083}}.

\bibitem{Andreo2009}
V.~Andreo, C.~Provensal, M.~Scavuzzo, M.~Lamfri, J.~Polop, Environmental
  factors and population fluctuations of {A}kodon azarae ({M}uridae:
  {S}igmodontinae) in central {A}rgentina, Austral Ecology 34~(2) (2009)
  132--142.
\newblock \href {http://dx.doi.org/10.1111/j.1442-9993.2008.01889.x}
  {\path{doi:10.1111/j.1442-9993.2008.01889.x}}.

\bibitem{Ra2012}
P.~K. Ra, M.~S. Nathawat, M.~Onagh, Application of multiple linear regression
  model through {GIS} and remote sensing for malaria mapping in {V}aranasi
  district, {I}ndia, Health Sci. J. 6~(4) (2012) 731--749.

\bibitem{Lopez2016}
M.~L\'opez-Ib\'a\~nez, J.~Dubois-Lacoste, L.~{P\'erez C\'aceres}, M.~Birattari,
  T.~St\"utzle, The irace package: Iterated racing for automatic algorithm
  configuration, Operations Research Perspectives 3 (2016) 43--58.

\end{thebibliography}

\end{document}